\documentclass{llncs}
\usepackage[T1]{fontenc}
\usepackage{graphicx}
\usepackage{subcaption}
\usepackage[utf8x]{inputenc}
\usepackage[english]{babel}
\usepackage{amssymb}
\usepackage{amsmath}

\graphicspath{ {./img/} }

\def\keywordname{{\bf Keywords:}}%
\begin{document}
\title{Correcting sampling biases via importance reweighting for spatial modeling}
%
%
\author{Prokhorov Boris \inst{3, 1, 2},
Diana Koldasbayeva \inst{3}, 
Zaytsev Alexey \inst{3}}
%
%
\institute{IITP RAS\and MIPT \and
Skoltech
\email{a.zaytsev@skoltech.ru}\\
}
\maketitle              
\begin{abstract}
In machine learning models, the estimation of errors is often complex due to distribution bias, particularly in spatial data such as those found in environmental studies. We introduce an approach based on the ideas of importance sampling to obtain an unbiased estimate of the target error. By taking into account difference between desirable error and available data, our method reweights errors at each sample point and neutralizes the shift. Importance sampling technique and kernel density estimation were used for reweighteing.

We validate the effectiveness of our approach using  artificial data that resemble real-world spatial datasets. 
Our findings demonstrate advantages of the proposed approach for the estimation of the target error, offering a solution to a distribution shift problem. Overall error of predictions dropped from 7\% to just 2\% and it gets smaller for larger samples. 

\keywords{Importance sampling, spatial modeling, Gaussian mixture, model validation.}
\end{abstract}

\section{Introduction}


In the rapidly advancing field of machine learning (ML), the accuracy and reliability of models are paramount. However, the path to improving these models faces a major hurdle: the issue of distributional bias. This bias leads to a skewed estimation of errors, consequently obstructing the assessment of the error in ML models. 

The problem is further magnified when dealing with spatial data, where the relationship between variables can be intricate and non-linear, and available observations are far from (usually uniform) target distribution.   
We can observe this problem in spatio-temporal data related to measurements at particular points~\cite{grabar2022predicting,barabesi2012spatial} or aggregation of events such as earthquakes~\cite{kail2021recurrent,gitis2015adaptive}.

In certain fields, particularly in environmental modeling, this challenge is not merely theoretical but manifests itself in tangible ways~\cite{koldasbayeva2022large}. Several research studies illustrate how distribution bias can distort models, resulting in erroneous predictions and potentially misguided decisions~\cite{robinson2018correcting}. Such examples stress the need for methodologies that can deliver an accurate estimate of the target error, free from the influences of bias.

We present a novel approach that seeks to overcome the limitations posed by distribution bias. Our contributions are two-fold:
\begin{itemize}
    \item Unbiased Estimation of Target Error: By employing an approach inspired by importance sampling, we propose an unbiased estimate of the target error. This statistical technique allows us to re-weight the sampled data, effectively eliminating the bias and providing a more accurate representation of the underlying distribution.
    \item Validation Using Artificial Data: To rigorously evaluate our approach, we utilize artificial data that closely mimic real-world spatial data used for ecological modeling. 
    Through this validation, we can assess the efficacy of our proposed method, demonstrating its robustness in accurately estimating the target error even in the presence of complex spatial distributions for available data.
\end{itemize}

\section{Related works}

Spatial sampling techniques have evolved significantly, creating practical datasets for real-world spatial analysis\cite{petrovskaia2021maxvol,wadoux2019clhs,barabesi2012spatial}. Our research addresses the challenges of working with arbitrary spatial samples. In this context, error estimation becomes a critical concern, and spatial cross-validation emerges as a key tool. It allows us to assess the predictive accuracy of spatial models across diverse spatial locations.

 

Customization of cross-validation is necessary to handle specific data structures effectively. One of these challenges is spatial autocorrelation (SAC), which introduces bias by linking measurements from nearby spatial points. Spatial cross-validation helps mitigate SAC's influence by dividing data into training and validation sets. This division reduces bias and significantly improves the reliability of model results.


Comparative studies highlight the superior performance of spatial cross-validation over traditional random cross-validation approaches. For instance, the research study~\cite{roberts2017cross} demonstrated through simulations and case studies that spatial cross-validation consistently outperforms random cross-validation, particularly when predicting new data, predictor space, or selecting causal predictors. Another research~\cite{ploton2020spatial} underscored the significance of acknowledging SAC's influence, showcasing that neglecting SAC can lead to overoptimistic assessments of model predictive power, emphasizing the need for accurate ecological mapping on a larger scale. Notably, the study~\cite{schratz2019hyperparameter} exhibited substantial performance differences, with up to 47 percent disparities, between bias-reduced spatial cross-validation and overoptimistic random cross-validation settings. This further accentuates the imperative to account for SAC's influence in achieving accurate model evaluations. The integration of spatial cross-validation and an awareness of SAC play a pivotal role in enhancing the credibility and robustness of model assessments across various domains, underlining the necessity of tailored cross-validation techniques when dealing with intricate spatial data structures.

To address distribution bias in spatial modeling, optimizing sampling designs is essential. In a relevant study~\cite{wadoux2019clhs}, researchers focused on soil mapping with random forest. They found that minimizing mean squared prediction error (MSE) through optimized sampling significantly enhanced accuracy, particularly for smaller samples. However, this approach relies on known soil values across all locations and is suited for subsampling existing datasets. For larger samples, a uniform spread in feature space is recommended. This highlights the sensitivity of comparing sampling strategies, especially with limited validation data.

An alternative approach is to define the range of applicability of a model~\cite{meyer2021predicting}.
However, it requires to design an uncertainty estimation procedure that is hard to do for all existing classes of methods for machine learning including linear models~\cite{erlygin2023uncertainty}, gradient boosting~\cite{malinin2020uncertainty}, and neural networks~\cite{kostenok2023uncertainty}.

Being introduced for estimation of statistics for complex distributions~\cite{bishop2006pattern},
importance sampling addresses the bias introduced by data distribution~\cite{tokdar2010importance}, while spatial cross-validation addresses the bias introduced by SAC.
Importance sampling principles could be applied to enhance spatial cross-validation by mitigating the bias introduced by spatial dependencies and ensuring more accurate assessments of spatial models.

We note, that existing methods to fight the distribution bias in error estimation are empirical.
Thus, there are no guarantee that they will reduce the error.
In contract, a properly developed principled approach would be a good and universal solution that doesn't require tuning hard-to-guess values of hyperparameters.





\section{Methods}

\subsection{Importance Sampling}

\subsubsection{Target error}

One can measure an error of a model at point $x$
\[ 
e(x) = (f(x) - \hat{f}(x))^2
\]
where $f(x)$ is the true value of a function and $\hat{f}(x)$ is our estimate.

We are interested in the risk estimate $R(e, p)$ for $x \sim p(x)$ in an area $X \subset \mathbb{R}^d$:
\[ 
R(e, p) = \int_{X} e(x) p(x) dx.
\]
A natural unbiased estimate of the $R(e, p)$ is the Monte-Carlo estimate
\begin{equation} 
\label{eq:true_p_estimate}
\hat{R}(e, p) = \frac1n \sum_{i = 1}^n e(x_i), x_i \sim p(x).
\end{equation}
Here and for all distributions below we consider their constraint to $X$.

\subsubsection{Error bias}

In reality we don't have points $x_i$ from the distribution $p(x)$.
Instead we have points from another distribution $g(x)$.
So, we have the estimate of the form:
\[
\hat{R}(e, g) = \frac1n \sum_{i = 1}^n e(x_i), x_i \sim g(x).
\]

An unbiased estimate of $\hat{R}(e, p)$ is:
\begin{equation}   
\label{eq:true_g_is_estimate}
R_I(e, p, g) = \frac1n \sum_{i = 1}^n e(x_i) \frac{p(x_i)}{g(x_i)}, x_i \sim g(x). 
\end{equation}

The similar idea leads to the importance sampling procedure~\cite{bishop2006pattern}, and it is easy to see that the provided estimate is unbiased.
Really, 
\[
\mathbb{E} \frac1n \sum_{i = 1}^n e(x_i) \frac{p(x_i)}{q(x_i)} = \int_{X} e(x) \frac{p(x)}{g(x)} g(x) dx = \int_{X} e(x) p(x) dx = R(e, p). 
\]

If $x$ is low-dimensional (e.g. for spatial data), we can estimate $g(x)$ from data. As $p(x)$ we typically consider a uniform distribution.
So, we obtain a pretty accurate ratio $\frac{p(x)}{g(x)}$.
Consequently, we get an estimate $\hat{R}_I(e, p, g)$:
\begin{equation}
\label{eq:est_g_is_estimate}
\hat{R}_I(e, p, g) = \frac1n \sum_{i = 1}^n e(x_i) \frac{p(x_i)}{\hat{g}(x_i)}, x_i \sim g(x). 
\end{equation}

\subsection{Spatial Data}
\subsubsection{GMM}
Spatial datasets often exhibit complex and multi-modal distributions, where different regions of the space may exhibit different statistical patterns. It is also widely known that spatial observations usually form spikes (for example around cities) and such spikes are well modeled by Bell curves. In order to align with these two concepts, we adopted the following model.
Gaussian mixture model (GMM, for details see \cite{bishop2006pattern}) is composed of multiple weighted Gaussian curves and can effectively capture multi-modal nature of spatial distribution. By assuming that the dataset is a sample from a mixture of Gaussian distributions, GMM provides a means to represent the underlying spatial patterns and heterogeneity. Having that said, we further consider $g$ as Gaussian mixture model.

\subsubsection{Additional restrictions}
In our study we naturally assume that it is theoretically possible to estimate target error from given dataset. This implies there is no 'holes' with respect to $p$ - areas with sufficient positive $p$-measure and absence of sample points inside it. 

\section{Experiments}
\subsection{Configuration}

For validation of our approach we conducted several numerical experiments. 

\begin{figure}[htbp]
\centering
\includegraphics[width=0.7\textwidth]{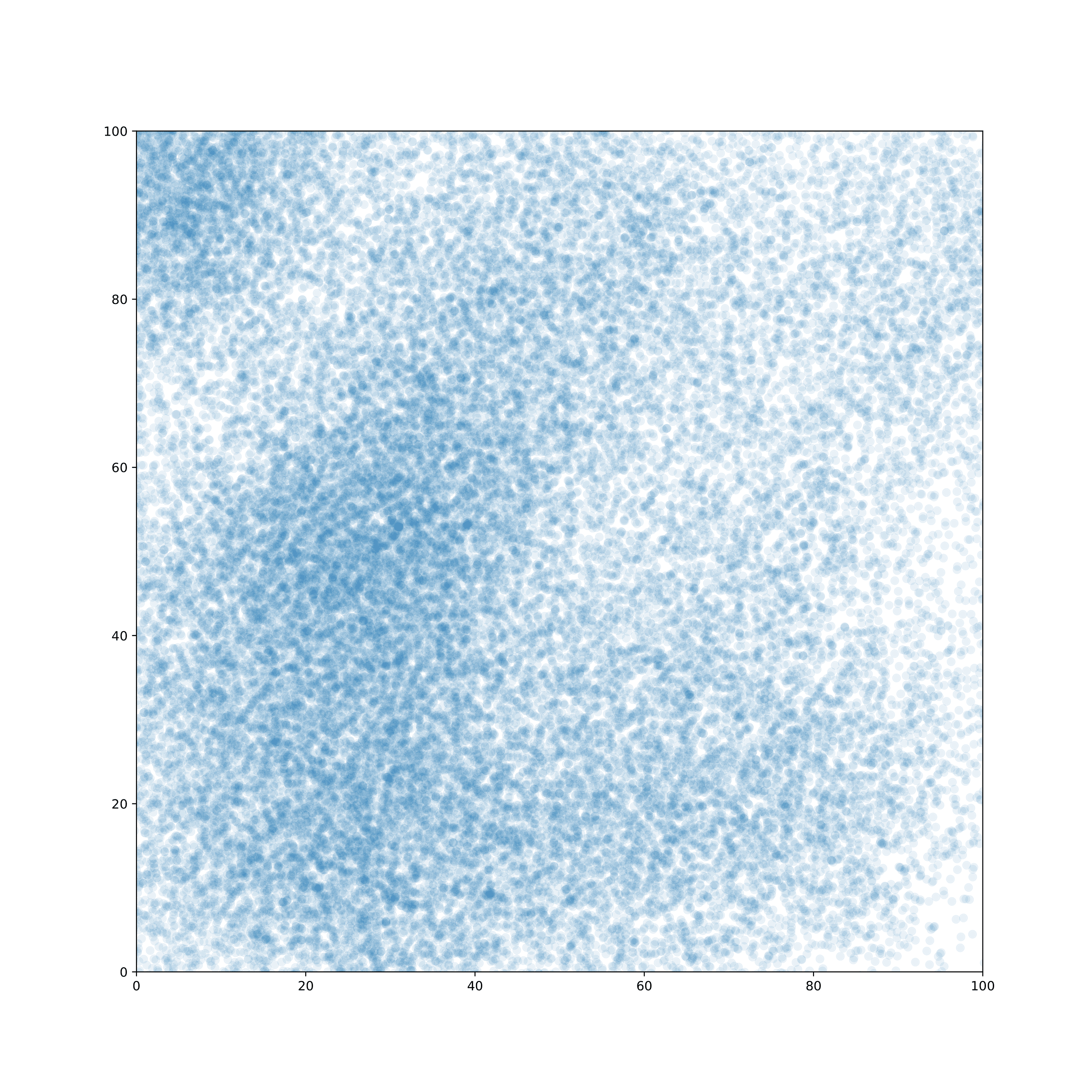}
\caption{GMM sample example for artificial two dimensional spatial data}
\label{fig:gmm_data_example}
\end{figure}

We used uniform distribution on the square $[0;100) \times [0;100)$ as $p$, as we expect that we are interested in estimation of the error of the uniform distribution of points.
For $g$ we used a Gaussian mixture. It consists of 20 randomly-centred components with each gaussian's covariance matrix chosen far from singular (it's eigenvalues $\geq$ 100) which assures the property of 'no holes' in the sample.
An example of generated GMM data is in Figure~\ref{fig:gmm_data_example}.
We selected these parameters of distribution to match a typical patter of the real data.

Since we are not interested in the accuracy of the model but in the accuracy of error estimate, the true function $f$ is a random linear or a random mixture of RBF (Gaussian functions).
We refer to them as Linear and GMM true functions correspondingly.
We consider an estimate $\hat{f}$ is either another random linear function or a Gradient boosting regression for the GMM case. 
Number of samples for estimation is $10 000$.  

We consider a baseline approach $MCE$ (Monte Carlo Error) and two variants of our $ISE$ (Importance Sampling Error) method.
$MCE$ approach correspond to the equation~\eqref{eq:true_p_estimate} in the assumption that $p$ and $g$ are close to each other universally adopted in the literature.
Basic $ISE$ assumes knowledge of the true distribution $g$, while for $ISE_g$ we estimate the density $g$ from data using KDE with Gaussian kernel~\cite{silverman2018density}.
For $ISE$ we use the equation~\eqref{eq:true_g_is_estimate}, 
and for $ISE_g$ we use the equation~\eqref{eq:est_g_is_estimate}.

\subsection{Validation procedure}

As a metric for real error estimation we used Mean Absolute Percentage Error (MAPE) and Root Mean Square Error (RMSE) scores relative to the true error based on 100 test-run results. Together they describe both relative and absolute accuracy of predictions.

\subsection{Main results}

Table~\ref{table:mape_relative} shows MAPE for estimation of the errors. 

We see, that our approach for the given experiment provides better estimation of the true error for both Linear and GMM true function.
The advantage of it is observed even in the case, when we use an estimation of $g$ instead of the true density.

\begin{table}[ht]
\caption{MAPE relative to real error in percents. The best value is highlighted in \textbf{bold} font.}
\label{table:mape_relative}
\centering
\begin{tabular}{cccc}
\hline
Function & $MCE$ (baseline) & $ISE$ & $ISE_e$\\
\hline
Linear & $6.2$ & $\mathbf{1.9}$ & $2$\\
GMM & $3.4$ & $\mathbf{1.5}$ &$ 2.3$ \\
\hline
\end{tabular}
\end{table}

\begin{figure}[ht]
\centering
\includegraphics[width=0.8\textwidth]{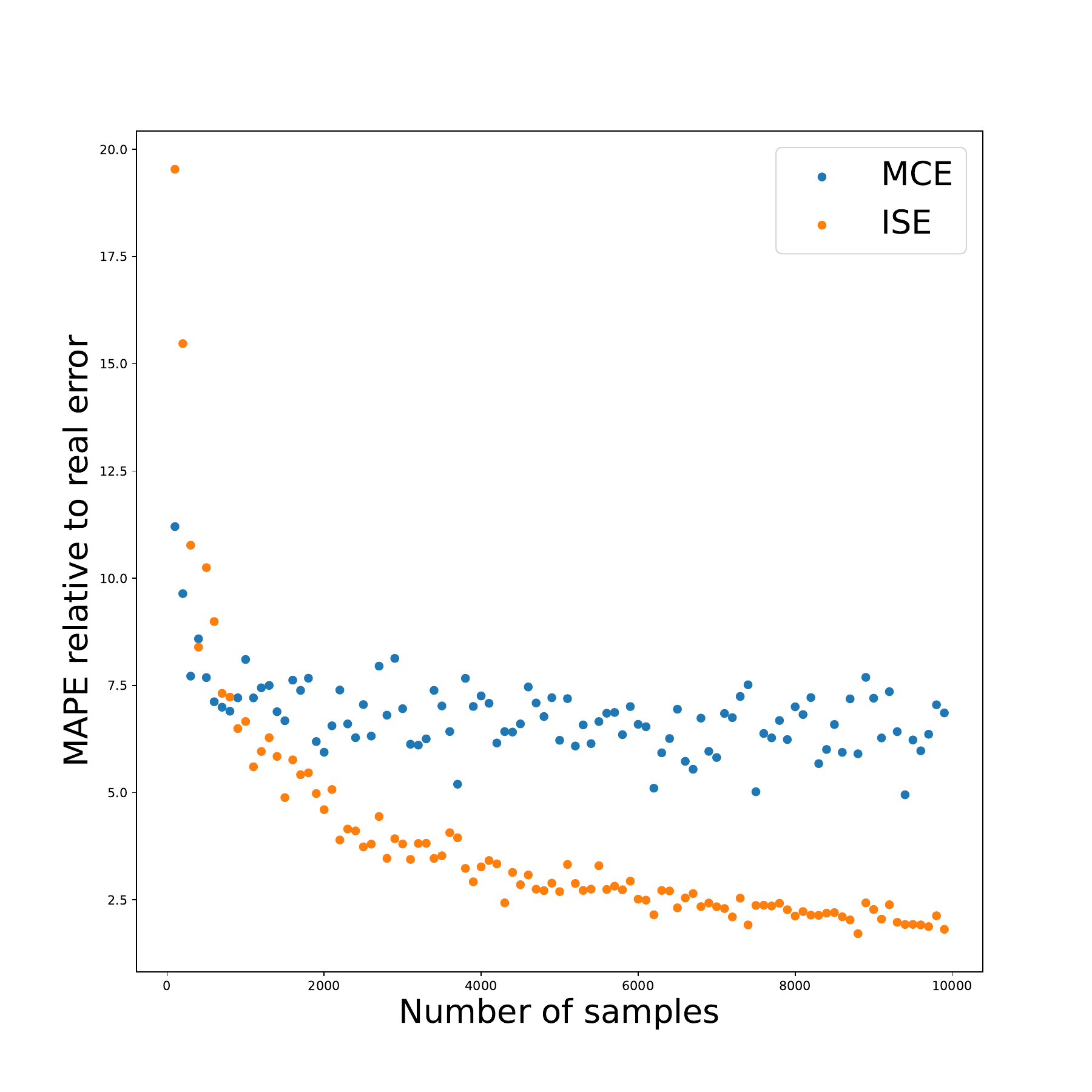}
\caption{$ISE$ vs $MCE$ with different sample sizes}
\label{fig:sample_size_dependence}
\end{figure}

\begin{figure}[ht]
    
    \centering
    \begin{subfigure}[b]{0.4\textwidth}
        \centering
        \includegraphics[width=\textwidth]{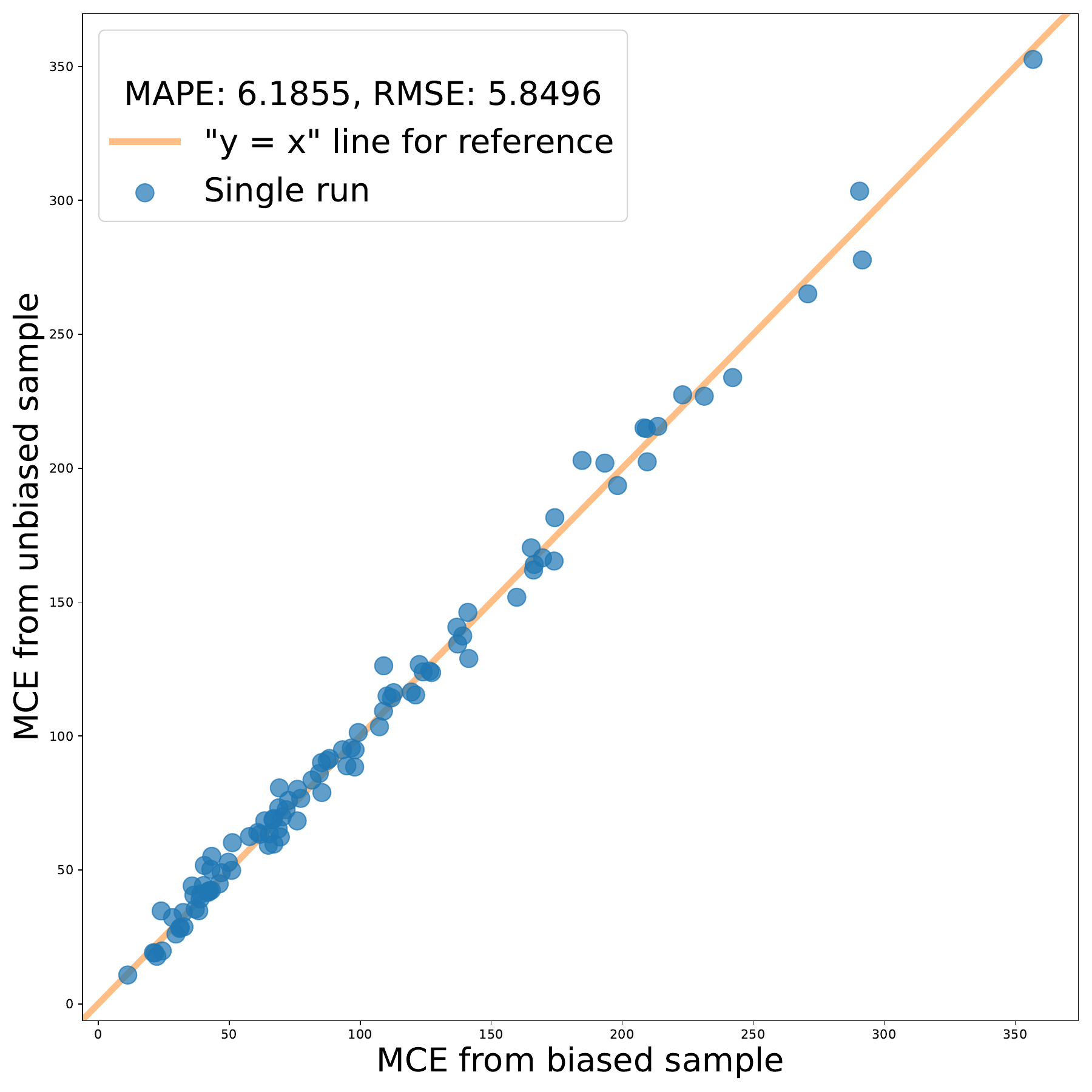}
        \caption{Linear function, $MCE$ estimator}
    \end{subfigure}
    \begin{subfigure}[b]{0.4\textwidth}
        \centering
        \includegraphics[width=\textwidth]{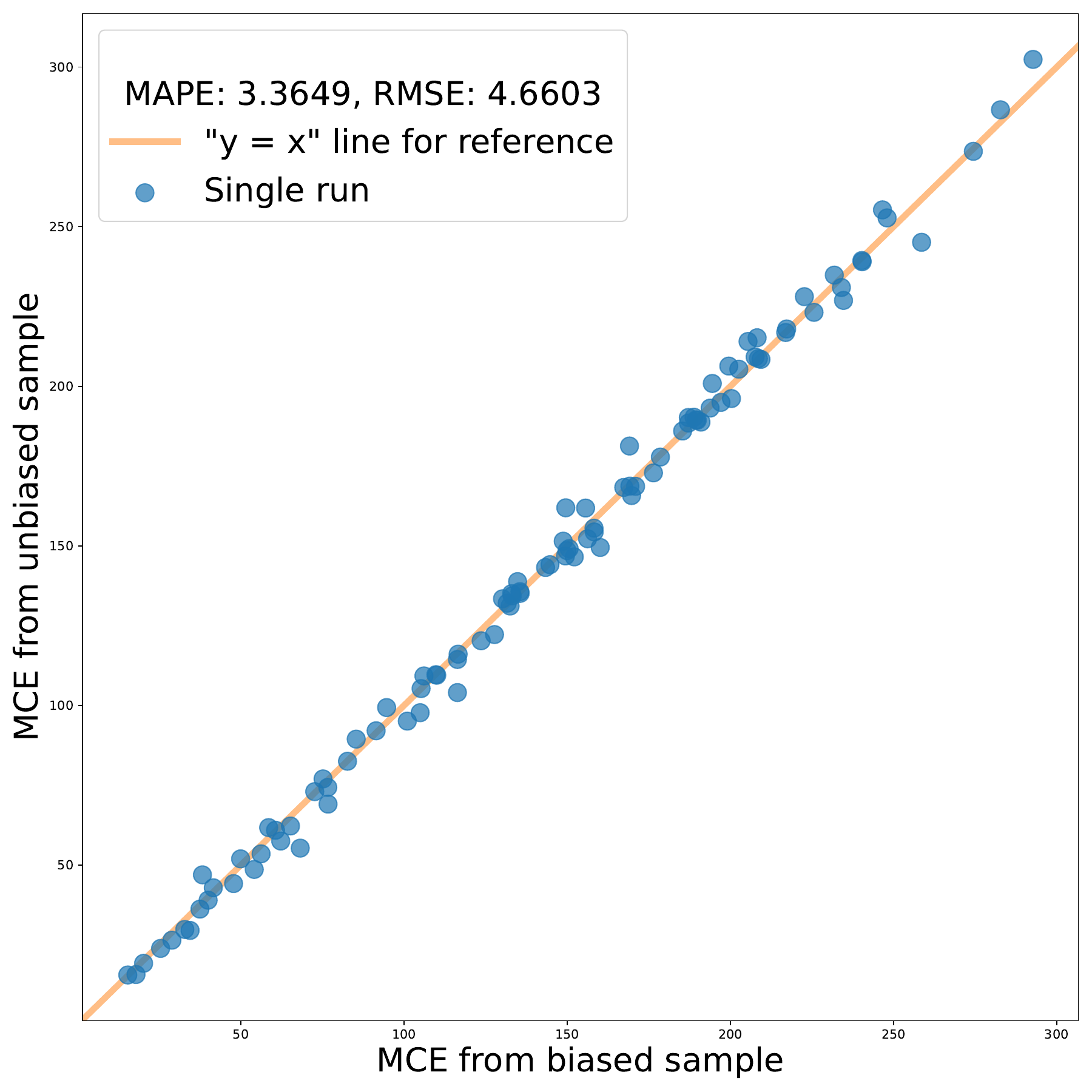}
        \caption{GMM function, $MCE$ estimator}
    \end{subfigure}
    \begin{subfigure}[b]{0.4\textwidth}
        \centering
        \includegraphics[width=\textwidth]{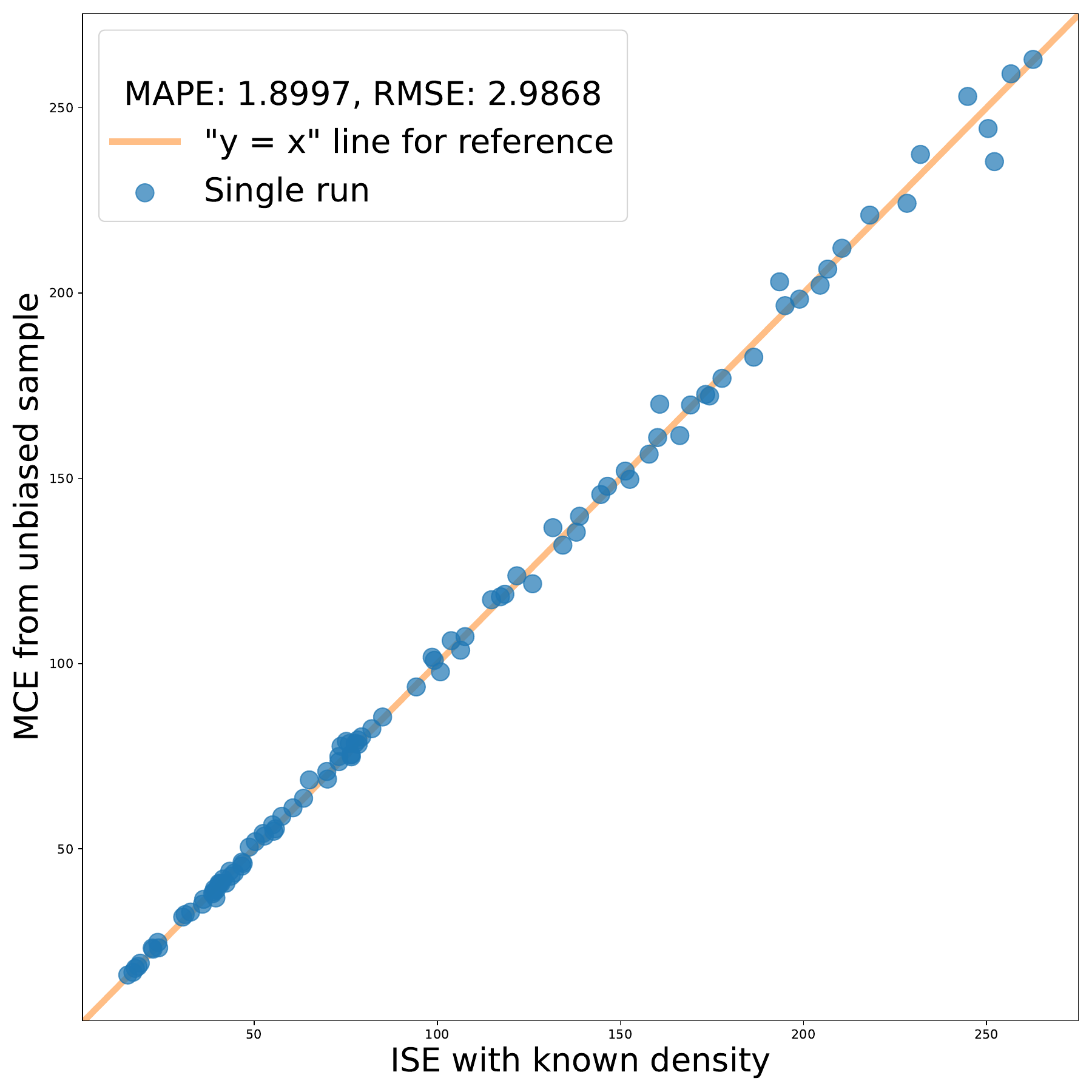}
        \caption{Linear function, $ISE$ estimator}
    \end{subfigure}
    \begin{subfigure}[b]{0.4\textwidth}
        \centering
        \includegraphics[width=\textwidth]{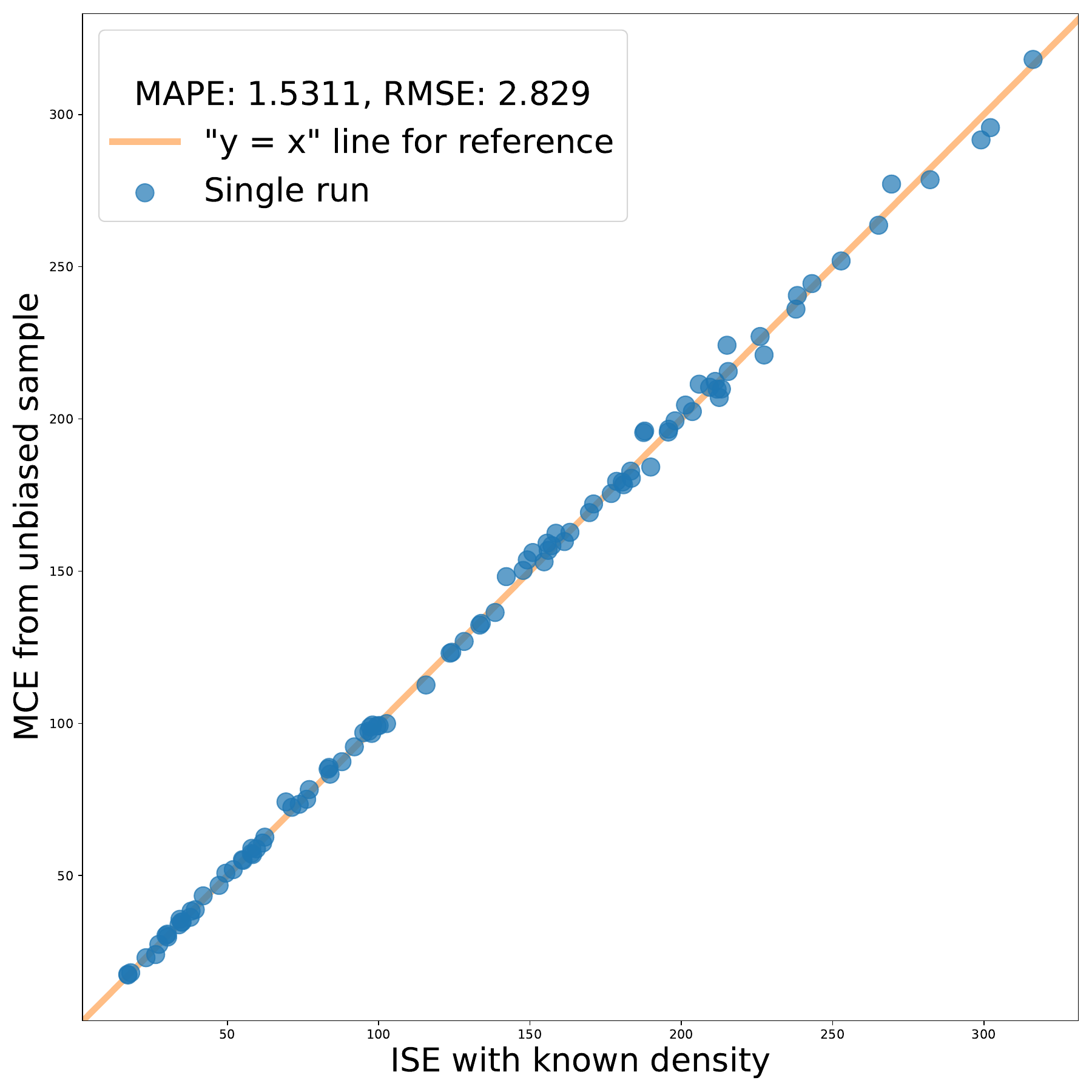}
        \caption{GMM function, $ISE$ estimator}
    \end{subfigure}
        \begin{subfigure}[b]{0.4\textwidth}
        \centering
        \includegraphics[width=\textwidth]{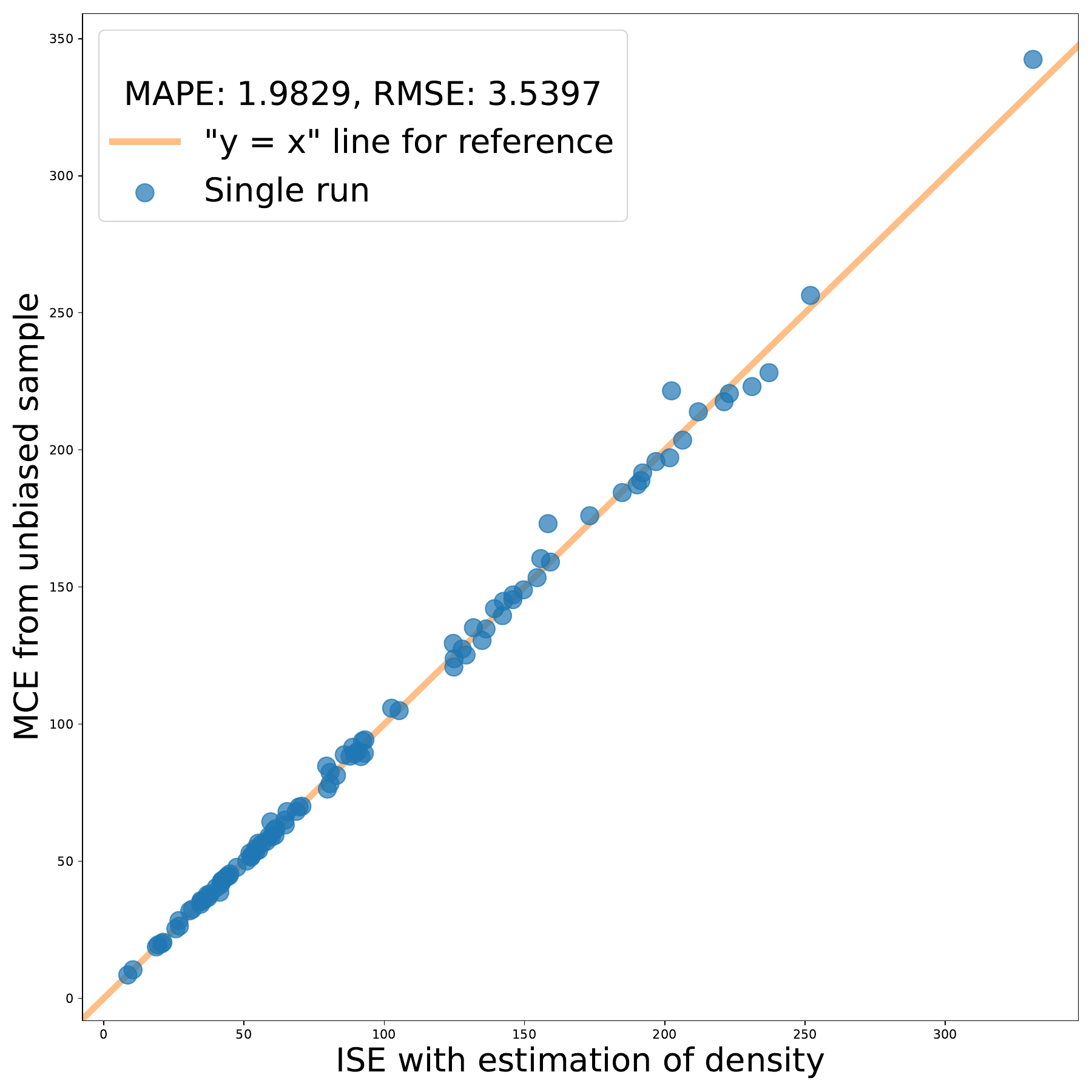}
        \caption{Linear function, $ISE_e$ estimator }
    \end{subfigure}
    \begin{subfigure}[b]{0.4\textwidth}
        \centering
        \includegraphics[width=\textwidth]{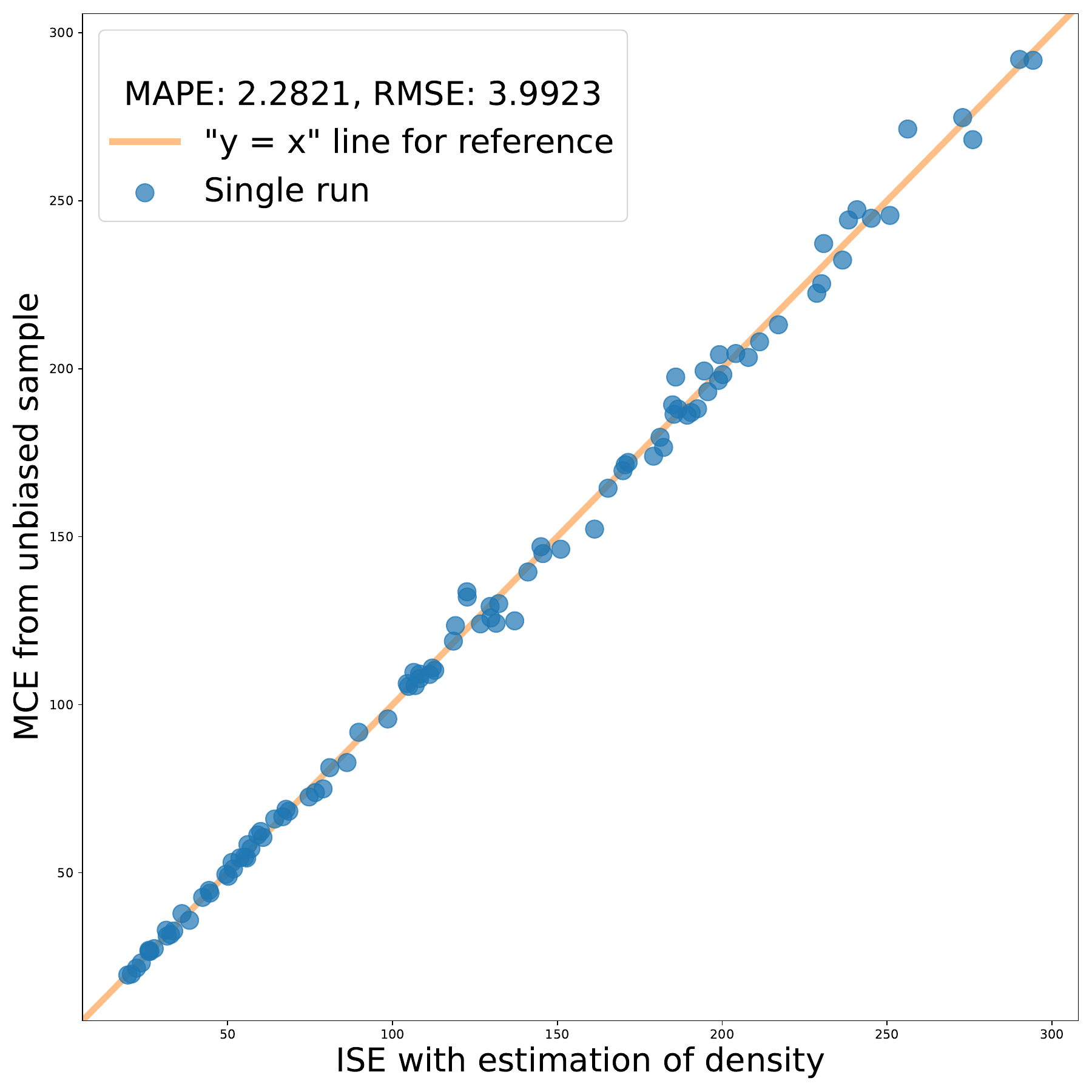}
        \caption{GMM function, $ISE_e$ estimator \phantom{aaa}}
    \end{subfigure}
    \caption{Comparison of considered approaches baseline $MCE$ and proposed $ISE$ and $ISE_e$ for different target functions linear (left) and Gaussian Mixture (GMM, right). Each point represents a single run.}
            \label{fig:scatter_plots}
\end{figure}

\subsection{Detailed analysis}

For detailed analysis we provide deeper investigation on the dependence of the quality of the error estimation for different number of samples and different target functions.

Figure~\ref{fig:sample_size_dependence} reveals that with growth of number of samples $MCE$ stays biased for about $0.07$ in average. Meanwhile $ISE$ in average gets closer and closer to real error, i.e. it is unbiased.
We also can note, that for extremely small sample sizes due to high variance $ISE$ gives worse estimations.

Figures~\ref{fig:scatter_plots} show low variance in the errors, providing strong evidence that Importance Sampling outperforms baseline, even if we use estimation of the density instead of the true density.

\subsection{Summary}
Experiments clearly show that $ISE$ and $ISE_g$ outperform $MCE$ under mild theoretical assumptions. We hope that it will also be true for real datasets, especially when sample size is increased. In any case, more research is required to assess applicability of the proposed method. Another uncovered part is the significance of precise error estimation. Experiments show that biased approach differed by only 7\% from the real error in average and impact of such miscalculation on the model training is unknown. Whether this impact is radical or not, one should keep in mind the possibility of distribution shift and subsequent model bias.

\section{Conclusions}

We considered the problem of error evaluation for geo-spatial data models and propose a method that provides unbiased estimates. 
By addressing the persistent issue of distribution bias, we open new avenues for improved modeling, particularly in fields where spatial data plays a central role.

Furthermore, the principles laid out in this study provide a foundation for future research, encouraging the development of even more refined techniques that can adapt and evolve in the face of ever-changing data landscapes.

\subsubsection*{Acknowledgements} 
This work was supported by the Russian Foundation for Basic Research grant 21-51-12005 NNIO\_a.

%
%
%
\bibliographystyle{splncs04}
\bibliography{main}

\end{document}